\documentclass{article} % For LaTeX2e
\usepackage{iclr2018_conference,times}
\usepackage{hyperref}
\usepackage{url}
\usepackage{booktabs}
\usepackage{natbib}

\usepackage{soul} % for highlighting
\usepackage{algorithm}
\usepackage{algorithmic}
\usepackage{amsmath}
\usepackage{amssymb}
\usepackage{mathtools}
\usepackage{pgf}
\usepackage{import}

\usepackage{microtype}      % microtypography
\usepackage{xfrac}

\newcommand\numberthis{\addtocounter{equation}{1}\tag{\theequation}}
\newcommand*\diff{\mathop{}\!\mathrm{d}}

\DeclareMathAlphabet{\mathcal}{OMS}{cmsy}{m}{n}

\DeclarePairedDelimiterX{\inner}[2]{\langle}{\rangle}{#1, #2}

\newcommand{\pdydx}[2]{ \frac{\partial #1}{\partial #2}}

\usepackage{nicefrac}
\usepackage{mdframed}
\usepackage{subcaption}
\usepackage{threeparttable, tablefootnote}

\title{The unreasonable effectiveness of the forget gate}

\author{Jos van der Westhuizen \\
University of Cambridge\\
\texttt{jv365@cam.ac.uk}
\And
Joan Lasenby \\
University of Cambridge\\
\texttt{jl221@cam.ac.uk}
}

\newcommand{\red}[1]{}
\newcommand{\todo}[1]{}
\iclrfinalcopy % Uncomment for camera-ready version, but NOT for submission.

\graphicspath{{figures/}}
\makeatletter
\def\input@path{{figures/}}
\makeatother

\begin{document}

\maketitle

\begin{abstract}
  Given the success of the gated recurrent unit, a natural question is whether all the gates of the long short-term memory (LSTM) network are necessary. Previous research has shown that the forget gate is one of the most important gates in the LSTM. Here we show that a forget-gate-only version of the LSTM with chrono-initialized biases, not only provides computational savings but outperforms the standard LSTM on multiple benchmark datasets and competes with some of the best contemporary models. Our proposed network, the JANET, achieves accuracies of 99\% and 92.5\% on the MNIST and pMNIST datasets, outperforming the standard LSTM which yields accuracies of 98.5\% and 91\%.
  
\end{abstract}

\section{Introduction}
Good engineers ensure that their designs are practical. After showing that a 
sequence analysis problem is best solved by the long short-term memory (LSTM) 
recurrent neural network, the next step is to devise an implementation enabling 
the often resource constrained real-world application. Given the success of the 
gated recurrent unit (GRU) \citep{ChoLearningPhraseRepresentations2014}, which 
uses two gates, the first approach to a more hardware efficient LSTM could be 
the elimination of redundant gates, if there are any. Because we seek a model 
more efficient than the GRU, only a single-gate LSTM model is a worthwhile 
endeavour. To motivate why this single gate should be the forget gate, we begin 
with the LSTM genesis.

In an era where training recurrent neural networks (RNNs) was notoriously 
difficult, \citet{hochreiter_long_1997} argued that having a single weight 
(edge) in the RNN to control whether input or output of a memory cell needs to 
be accepted or ignored, creates conflicting updates (gradients). Essentially, 
the long and short-range error act on the same weight at each step, and with 
sigmoid activated units, this results in the gradients vanishing faster than 
the weights can grow. They proceeded to propose the long short-term memory 
(LSTM) unit recurrent neural network, which had multiplicative input and output 
gates. These gates would mitigate the conflicting update issue by 
``protecting'' the cells from irrelevant information, either from the input or 
from the output of other cells.

This first version of the LSTM had only two gates; it was \citet{GersLearningForgetContinual2000} who realized that if there is no mechanism for the memory cells to forget information, they may grow indefinitely and eventually cause the network to break down. As a solution, they proposed another multiplicative gate for the LSTM architecture, known as the forget gate -- completing the version of the LSTM that we know today\footnote{It's interesting to note the difference between the motivations that lead to the LSTM and the chain-of-thought that yielded the gated recurrent unit (GRU). Cho was ``not well aware'' \citep[\S4.2.3]{ChoNaturalLanguageUnderstanding2015} of the LSTM when he, together with collaborators, designed the GRU. In contrast to the conflicting update problem \citep{hochreiter_long_1997} and the indefinite state growth \citep{GersLearningForgetContinual2000} arguments, \citet{ChoNaturalLanguageUnderstanding2015} approached the RNN problem by thinking of it as a computer processor with memory registers. In the case of computers, we do not want to overwrite all the registers (memory values) at each step. Therefore, the RNN requires an update gate, which controls the hidden states (registers) that are overwritten (the update gate in the GRU is akin to the combined function of the input and forget gates in the LSTM). Furthermore, we do not necessarily need to read all the registers at each time step, only the important ones. Thus another gate is required in the RNN (the reset gate) to regulate the registers considered. Ideally, all of the gating operations would be binary values, but such values would result in zero gradients. Fortunately, the sigmoid or tanh nonlinearities provide leaky versions of these gating mechanisms and have smooth gradients.}.

It wasn't until many years later that \citet{greff_lstm:_2015} and \citet{jozefowicz2015empirical} simultaneously discovered the forget gate to be the crucial ingredient of the LSTM. \citet{GersLearningForgetContinual2000} proposed initializing the forget gate biases to positive values and \citet{jozefowicz2015empirical} showed that an initial bias of 1 for the LSTM forget gate makes the LSTM as strong as the best of the explored architectural variants (including the GRU) \citep[\S10.10.2]{GoodfellowDeepLearning2016}. Given the new-found importance of the forget gate, would the input and output gates have been found necessary if the LSTM was conceived with only a forget gate? 

In this work, we take the liberty of exploring the gains introduced by the sole use of the forget gate. On the five tasks explored, use of only the forget gate provides a better solution than the use of all three LSTM gates. Many improvements have been proposed for the LSTM, which we review in the following section.

\section{Related work}
With some success, many studies have improved the LSTM by making the cell more complex \citep{NeilPhasedLSTMAccelerating2016,HeWiderDeeperCheaper2017,FraccaroSequentialNeuralModels2016a,KruegerZoneoutRegularizingRNNs2017,GravesPracticalVariationalInference2011}, with classic examples being peephole connections \citep{GersRecurrentnetsthat2000} and depth gated LSTMs \citep{YaoDepthgatedRecurrentNeural2015}. Similarly, several studies have proposed recurrent neural networks (RNN) simpler than the LSTM yet still competitive, such as the skip-connected RNN \citep{ZhangArchitecturalComplexityMeasures2016}, the unitary RNN \citep{arjovsky2016unitary}, the Delta-RNN \citep{OrorbiaIILearningSimplerLanguage2017}, and the identity RNN \citep{le_simple_2015}. However, one of the most thorough studies on the architecture of the LSTM is probably the study by \citet{greff_lstm:_2015} (5,400 experiment simulations). They explored the following LSTM variants individually:
\begin{itemize}%[noitemsep]
	\item No input gate
	\item No forget gate
	\item No output gate
	\item No input activation function
	\item No output activation function
	\item No peepholes
	\item Coupled input and forget gate
	\item Full gate recurrence
\end{itemize}
The first five variants are self-explanatory. Peepholes 
\citep{GersRecurrentnetsthat2000} connect the cell to the gates, adding an 
extra term to the pre-activations of the input, output, and forget gates. The 
coupled input and forget gate variant uses only one gate for modulating the 
input and the cell recurrent self-connections, i.e., $ \textbf{f} = 
\textbf{1}-\textbf{i} $. Full gate recurrence is the initial setup of 
\citet{hochreiter_long_1997}, wherein all the gates receive recurrent inputs 
from all gates at the previous time step. This cumbersome architecture requires 
9 additional recurrent weight matrices and did not feature in any of their 
later papers. Interestingly, the results in \citet{greff_lstm:_2015} indicate 
that none of the variants significantly improve on the standard LSTM. The 
forget gate was found to be essential, but a forget-gate-only variant was not 
explored.

Two studies that are closely related to ours are those by 
\citet{ZhouMinimalgatedunit2016} and \citet{WuInvestigatinggatedrecurrent2016}. 
The former successfully implemented a similar gate reduction to the gated 
recurrent unit (GRU); they couple the reset (input) gate to the update (forget) 
gate and show that this minimal gated unit (MGU) achieves a performance similar 
to the standard GRU with only two-thirds of the parameters. The study by 
\citet{WuInvestigatinggatedrecurrent2016} proposes a gate reduction similar to 
that of ours for LSTMs. They demonstrate that their {\em simple} LSTM achieves 
the same performance as the standard LSTM on a speech synthesis task. Compared 
with our work, they keep the hyperbolic tangent activation function on the 
memory cell, and their implementation did not employ the same bias 
initialization scheme, which we show is paramount for successful implementation 
of these models over a wide range of datasets. We became aware of 
these studies after having completed most of our work; our simplification of 
the LSTM provides a network that yields classification accuracies at least as 
good as the standard LSTM and often performs substantially better -- a result 
not achieved by the models proposed in the afore-mentioned studies.\todo{possibly cite the light-gru paper that was published at the same time.}

\section{Just Another NETwork}\label{sec:janet}
Recurrent neural networks (RNNs) typically create a lossy summary $\textbf{h}_T $ of a sequence. It is lossy because it maps an arbitrarily long sequence $ \textbf{x}_{1:T} $ into a fixed length vector. As mentioned before, recent work has shown that this forgetting property of LSTMs is one of the most important \citep{greff_lstm:_2015,jozefowicz2015empirical}. Hence, we propose a simple transformation of the LSTM that leaves it with only a forget gate, and since this is Just Another NETwork (JANET), we name it accordingly. We start from the standard LSTM \citep{lipton_critical_2015}, which, with symbols taking their standard meaning, is defined as
\begin{align*}
\textbf{i}_t &= \sigma(\textbf{U}_{i}\textbf{h}_{t-1} + 
\textbf{W}_{i}\textbf{x}_t + \textbf{b}_{i})\\
\textbf{o}_t &= \sigma(\textbf{U}_{o}\textbf{h}_{t-1} + 
\textbf{W}_{o}\textbf{x}_t + \textbf{b}_{o})\\
\textbf{f}_t &= \sigma(\textbf{U}_{f}\textbf{h}_{t-1} + 
\textbf{W}_{f}\textbf{x}_t + \textbf{b}_{f})\\
\textbf{c}_t &= \textbf{f}_t\odot \textbf{c}_{t-1}+\textbf{i}_t\odot 
\tanh(\textbf{U}_{c}\textbf{h}_{t-1} + \textbf{W}_{c}\textbf{x}_t + 
\textbf{b}_{c})\\
\textbf{h}_t &= \textbf{o}_t \odot \tanh(\textbf{c}_t). \numberthis 
\label{eq:lstm_cell_f}
\end{align*}
To transform the above into the JANET architecture, the input and output gates are removed. It seems sensible to have the accumulation and deletion of information be related, therefore we couple the input and forget modulation as in \citet{greff_lstm:_2015}, which is similar to the leaky unit implementation \citep[\S8.1]{jaeger2002tutorial}. Furthermore, the $ \tanh $ activation of $ \textbf{h}_t $ shrinks the gradients during backpropagation, which could exacerbate the vanishing gradient problem, and since the weights $ \textbf{U}_* $ can accommodate values beyond the range [-1,1], we can remove this unnecessary, potentially problematic, $ \tanh $ nonlinearity. The resulting JANET is given by 
\begin{align*}
\textbf{f}_t &= \sigma(\textbf{U}_{f}\textbf{h}_{t-1} + 
\textbf{W}_{f}\textbf{x}_t + \textbf{b}_{f})\\
\textbf{c}_t &= \textbf{f}_t\odot 
\textbf{c}_{t-1}+(\textbf{1}-\textbf{f}_t)\odot 
\tanh(\textbf{U}_{c}\textbf{h}_{t-1} + \textbf{W}_{c}\textbf{x}_t + 
\textbf{b}_{c}) \\
\textbf{h}_t &= \textbf{c}_t. \numberthis \label{eq:janet}
\end{align*}
Intuitively, allowing slightly more information to accumulate than the amount forgotten would make sequence analysis easier. We found this to be true empirically by subtracting a pre-specified value $ \beta $ from the input control component\footnote{$ \beta $ is a constant-valued column vector of the appropriate size.}, as given by
\begin{align*}
	\textbf{s}_t &= \textbf{U}_{f}\textbf{h}_{t-1} + 
	\textbf{W}_{f}\textbf{x}_t + \textbf{b}_{f}\\
	\tilde{\textbf{c}}_t &= \tanh(\textbf{U}_{c}\textbf{h}_{t-1} + \textbf{W}_{c}\textbf{x}_t + 
	\textbf{b}_{c}) \\
	\textbf{c}_t &= \sigma(\textbf{s}_t)\odot 
	\textbf{c}_{t-1}+(\textbf{1}-\sigma(\textbf{s}_t-\beta))\odot \tilde{\textbf{c}}_t \\	
	\textbf{h}_t &= \textbf{c}_t. \numberthis \label{eq:janet2}
\end{align*}
We speculate that the value for $ \beta $ is dataset dependent, however, we found that setting $ \beta=1 $ provides the best results for the datasets analysed in this study, which have sequence lengths varying from 200 to 784.

If we follow the standard parameter initialization scheme for LSTMs, the JANET 
quickly encounters a problem. The standard procedure is to initialize the 
weights $ \textbf{U}_* $ and $ \textbf{W}_* $ to be distributed as $ 
\mathcal{U}\left[-\nicefrac{\sqrt{6}}{\sqrt{n_l+n_{l+1}}},\nicefrac{\sqrt{6}}{\sqrt{n_l+n_{l+1}}}\right]
 $, where $ n_l $ is the size of each layer $ l $ 
\citep{HeDelvingDeepRectifiers2015,GlorotUnderstandingdifficultytraining2010}, 
and to initialize all biases to zero except for the forget gate bias $ 
\textbf{b}_f $, which is initialized to one \citep{jozefowicz2015empirical}. 
Hence, if the values of both input and hidden layers are zero-centred over 
time, $ \textbf{f}_t $ will be centred around $ \sigma(1)=0.7311$. In this case, 
the memory values $ \textbf{c}_t $ of the JANET would not be retained for more 
than a couple of time steps. This problem is best exemplified by the MNIST 
dataset \citep{lecun_mnist_1998} processed in scanline order 
\citep{CooijmansRecurrentBatchNormalization2016}; each training example contains many 
consecutive zero-valued subsequences, each of length 10 to 20. In the best case 
scenario -- a length 10 zero-valued subsequence -- the memory values at the end 
of the subsequence would be centred around 
\begin{equation}\label{eq:mem}
\textbf{c}_{t+10}=\textbf{f}_t^{10}\odot\textbf{c}_t\leq0.7311^{10}\textbf{c}_t\leq0.04363\textbf{c}_t.
\end{equation}
Thus, with the standard initialization scheme, little information would be propagated during the forward pass and in turn, the gradients will quickly vanish.

Fortunately, the recent work by \citet{TallecCanrecurrentneural2018} proposed a more suitable initialization scheme for the forget gate biases of the LSTM. To motivate this initialization scheme we start by re-writing the leaky RNN \citep[\S8.1]{jaeger2002tutorial}
\begin{equation}\label{eq:leaky}
\textbf{h}_{t+1}=\alpha\odot \tanh(\textbf{U}\textbf{h}_{t}+\textbf{W}\textbf{x}_t+\textbf{b})+(\textbf{1}-\alpha)\odot\textbf{h}_{t},
\end{equation}
as its continuous time version, by making use of the first order Taylor expansion $ h(t+\delta t) \approx h(t) + \delta t \frac{\diff h(t)}{\diff t} $ and a discretization step $ \delta t =1 $,
\begin{equation}\label{eq:cont_leaky}
\frac{\diff \textbf{h}(t)}{\diff t} = \alpha \odot \tanh \Big(\textbf{U}\textbf{h}(t)+\textbf{W}\textbf{x}(t)+\textbf{b}\Big) - \alpha \odot \textbf{h}(t).
\end{equation}
\citet{TallecCanrecurrentneural2018} state that in the free regime, when inputs stop after a certain time $ t_0,\ x(t) = 0  $ for $ t>t_0 $, with $ \textbf{b}=0 $ and $ \textbf{U}=0 $, eq. \ref{eq:cont_leaky} becomes
\begin{align*}
\frac{\diff \textbf{h}(t)}{\diff t} &= -\alpha \textbf{h}(t)\\
\int_{t_0}^{t}\frac{1}{\textbf{h}(t)}\diff \textbf{h}(t) &= -\alpha\int_{t_0}^{t}\diff t\\
\textbf{h}(t) &= \textbf{h}(t_0)\exp(-\alpha(t-t_0)).
\numberthis \label{eq:time_const}
\end{align*}
From eq. \ref{eq:time_const} the hidden state $ \textbf{h} $ will decrease to $ e^{-1} $ of its original value over a time proportional to $ 1/\alpha $. This $ 1/\alpha $ can be interpreted as the characteristic forgetting time, or the time constant, of the recurrent neural network. Therefore, when modelling sequential data believed to have dependencies in a range $[T_{\text{min}},T_{\text{max}}] $, it would be sensible to use a model with a forgetting time lying in approximately the same range, i.e., having $ \alpha \in [\frac{1}{T_{\text{max}}},\frac{1}{T_{\text{min}}}]^d  $, where $ d $ is the number of hidden units. 

For the LSTM, the input gate $ \textbf{i} $ and the forget gate $ \textbf{f} $ 
learn time-varying approximations of $ \alpha $ and $ (1-\alpha) $, 
respectively. Obtaining a forgetting time centred around $ T $ requires $ 
\textbf{i} $ to be centred around $ 1/T $ and $ \textbf{f} $ to be centred 
around $ (1-1/T) $. Assuming the shortest dependencies to be a single time 
step, \citet{TallecCanrecurrentneural2018} propose the \textbf{chrono 
initializer}, which initializes the LSTM gate biases as
\begin{align*}
\textbf{b}_f&\sim\log(\mathcal{U}[1,T_{\text{max}}-1]) \\
\textbf{b}_i&=-\textbf{b}_f, \numberthis
\end{align*}
with $ T_{\text{max}} $ the expected range of long-term dependencies and $ \mathcal{U} $ the uniform distribution. Importantly, these are only the initializations, and the gate biases are allowed to change independently during training.

Applying chrono initialization to the forget gate $ \textbf{f} $ of the JANET\footnote{The memory cell biases $ \textbf{b}_c $ are initialized to zero.}, mitigates the memory issue (eq. \ref{eq:mem}). With the values of the input and hidden layers zero-centred over time, the forget gate corresponding to a long-range ($ T_{\text{max}} $) cell will have an activation of
\begin{align*}
\sigma(\log(T_{\text{max}}-1)) =& \frac{1}{1+\exp(-\log(T_{\text{max}}-1))}\\ \xrightarrow[T_{\text{max}}\rightarrow\infty]{}& 1. \numberthis
\end{align*}
Consequently, for the MNIST memory problem ($ T_{\text{max}}-1=783 $), these long-range cells would retain most of their information, even after 20 consecutive zeros
\begin{align*}
f^{long} &= \frac{1}{1+\exp(-\log(783))} \geq 0.9987\\
c_{t+20}^{long}&=(f^{long})^{20}c_t^{long}\geq 0.9987^{20}c_t^{long}\geq0.97c_t^{long}. \numberthis \label{eq:long_cell}
\end{align*}

For the JANET, chrono initialization provides an elegant implementation of skip-like connections between the memory cells over time. It has long been known that skip connections mitigate the vanishing gradient problem \citep{SrivastavaTrainingVeryDeep2015a,LinLearninglongtermdependencies1996}. A systematic study of recurrent neural networks (RNNs) by \citet{ZhangArchitecturalComplexityMeasures2016} found that explicitly adding skip connections in the RNN graph improves performance by allowing information to be transmitted directly between non-consecutive time steps. For RNNs, they devise the recurrent skip coefficient, a value that measures the number of time steps through which unimpeded flow of information is allowed, and argue that higher values are usually better. Furthermore, skip connections are responsible for much of the boom in machine learning; they are the pith of the powerful residual networks \citep{HeDeepResidualLearning2015}, highway networks \citep{SrivastavaTrainingVeryDeep2015a}, and the WaveNet \citep{VanDenOordWaveNetGenerativeModel2016}. A natural question that follows, is how the skip-connections influence the gradients of the JANET and the LSTM.

\subsection{A comparison of gradients}
Before comparing the gradients of the LSTM and the JANET we provide some preliminaries. We denote the derivatives of the element-wise nonlinearities by the following:
\begin{align*}
&\sigma'(x)=\sigma(x)(1-\sigma(x)), 
&0<\sigma'(x)\leq 0.25\\
&\tanh'(x)=1-\tanh^2(x), 
&0<\tanh'(x)\leq1
\numberthis
\end{align*}
For brevity, we denote the pre-activation vectors in eq. \ref{eq:lstm_cell_f} and \ref{eq:janet} as
\begin{equation}
\textbf{s}_{i,o,f,c} = \textbf{U}_{i,o,f,c}\textbf{h}_{t} + \textbf{W}_{i,o,f,c}\textbf{x}_{t+1} + \textbf{b}_{i,o,f,c}.
\end{equation}
Lastly, we consider a diagonal matrix as a vector of its diagonal elements. 
Thus, a derivative of an element-wise multiplication of two vectors is written 
as a vector. Consider the following derivative of an element-wise 
multiplication of vectors $ \{\textbf{a},\textbf{b}\}\in\mathcal{R}^3 $
\begin{align*}
\pdydx{\textbf{v}}{\textbf{a}} &= 
\pdydx{}{\textbf{a}}\textbf{b}\odot\textbf{a}\\
&=	\begin{bmatrix}
\pdydx{v_1}{a_1} & \pdydx{v_1}{a_2} & \pdydx{v_1}{a_3} \\
\pdydx{v_2}{a_1} & \pdydx{v_2}{a_2} & \pdydx{v_2}{a_3} \\
\pdydx{v_3}{a_1} & \pdydx{v_3}{a_2} & \pdydx{v_3}{a_3} \\
\end{bmatrix}\\
&=	\begin{bmatrix}
b_1 & 0 & 0 \\
0 & b_2 & 0 \\
0 & 0 & b_3 \\
\end{bmatrix}, \numberthis
\end{align*}
which we write as
\begin{equation}
\pdydx{\textbf{v}}{\textbf{a}} = \textbf{b}.
\end{equation}

Here we compare the gradient propagation through the memory cells of a single-layer JANET with that of a single-layer LSTM. To analyse this flow of information we can compute the gradient $ \nicefrac{\partial J}{\partial \textbf{c}_{t}} $ of the objective function $ J $ with respect to some arbitrary memory vector $ \textbf{c}_t $. Starting with the JANET (eq. \ref{eq:janet}), we re-write it as
\begin{align*}
	\textbf{f}_{t+1} &= \sigma(\textbf{s}_f)\\
	\textbf{c}_{t+1} &= \textbf{f}_{t+1}\odot\textbf{c}_{t}
	+(\textbf{1}-\textbf{f}_{t+1})\odot \tanh(\textbf{s}_c). \numberthis
\end{align*}
For this architecture the gradient of the objective function is given by
\begin{equation}\label{eq:janet_prod_deriv}
\pdydx{J}{\textbf{c}_{t}}=\pdydx{J}{\textbf{c}_T}\prod_{k=t}^{T-1}\left[\pdydx{\textbf{c}_{k+1}}{\textbf{c}_k}\right],
\end{equation}
with
\begin{align*}
	\pdydx{\textbf{c}_{t+1}}{\textbf{c}_t}&= \textbf{U}_f\sigma'(\textbf{s}_f)\odot\textbf{c}_{t}
	+\sigma(\textbf{s}_f) 
	+(1-\sigma(\textbf{s}_f))\odot(\textbf{U}_c\tanh'(\textbf{U}_c\textbf{c}_t))
	\\
	&\quad -\sigma'(\textbf{s}_f)\odot(\textbf{U}_f\tanh(\textbf{U}_c\textbf{c}_t)).
	\numberthis \label{eq:janet_deriv}
\end{align*}
Assuming that the input and hidden layers are zero-centred over time (as for the memory problem eq. \ref{eq:mem}) and all the forget gate biases are initialized to the longest range (eq. \ref{eq:long_cell}), $ \sigma(\textbf{s}_f)$ will typically take values of one\footnote{With the biases large enough for $ \sigma(\textbf{s}_f) \approx \sigma(\textbf{s}_f-\beta) $} and $ \sigma'(\textbf{s}_f) $ values near zero. In this scenario, we see that all but one of the terms in eq. \ref{eq:janet_deriv} reduce to zero and we have
\begin{equation}\label{eq:skip-con}
\pdydx{\textbf{c}_{t+1}}{\textbf{c}_t} \approx \textbf{1},
\end{equation}
meaning that gradients from distant memory cells $ \textbf{c}_t $ are largely unaffected by the sequence length.

Moving on to the LSTM, we re-write eq. \ref{eq:lstm_cell_f} as
\begin{align*}
	\textbf{i}_{t+1},\textbf{o}_{t+1},\textbf{f}_{t+1} &= \sigma(\textbf{s}_{i,o,f})\\
	\textbf{c}_{t+1} &= \textbf{f}_{t+1}\odot \textbf{c}_{t}+\textbf{i}_{t+1}\odot \tanh(\textbf{s}_c)\\
	\textbf{h}_{t+1} &= \textbf{o}_{t+1} \odot \tanh(\textbf{c}_{t+1}). \numberthis
\end{align*}
Here the gradient of the objective function is
\begin{align*}
	\pdydx{J}{\textbf{c}_{t}}=\pdydx{J}{\textbf{h}_t}\pdydx{\textbf{h}_t}{\textbf{c}_t} + \pdydx{J}{\textbf{c}_{t+1}}\pdydx{\textbf{c}_{t+1}}{\textbf{c}_{t}}
	=\pdydx{J}{\textbf{h}_t}\pdydx{\textbf{h}_t}{\textbf{c}_t} + \pdydx{J}{\textbf{c}_{t+1}}\textbf{f}_{t+1}. \numberthis \label{eq:lstm_deriv2}
\end{align*}
With a forget gate chrono-initialized to a hypothetical value of one and with $ \pdydx{J}{\textbf{h}_t}\pdydx{\textbf{h}_t}{\textbf{c}_t}=0 $, the LSTM would permit unhindered gradient propagation. Under standard and chrono-initialization schemes, however, this $ \pdydx{J}{\textbf{h}_t}\pdydx{\textbf{h}_t}{\textbf{c}_t} $ term is unlikely to be zero. First,
\begin{equation}
\pdydx{\textbf{h}_t}{\textbf{c}_t} = \textbf{o}_{t}\odot\tanh'(\textbf{c}_t),
\end{equation}
which is non-zero with $ 0\leq\textbf{o}_{t}\leq1 $ (centred around 0.5 under the memory problem assumptions eq. \ref{eq:mem}) and $ 0\leq \tanh'(\textbf{c}_t)\leq1 $. Second,
\begin{equation}
\pdydx{J}{\textbf{h}_t} = \pdydx{J}{\textbf{h}_{t+1}}\pdydx{\textbf{h}_{t+1}}{\textbf{h}_{t}} + \pdydx{J}{\textbf{h}_{t+1}}\pdydx{\textbf{h}_{t+1}}{\textbf{c}_{t+1}}\pdydx{\textbf{c}_{t+1}}{\textbf{h}_{t}},
\end{equation}
where under chrono-initialized assumptions
\begin{align*}
	\pdydx{\textbf{c}_{t+1}}{\textbf{h}_t} = 
	\textbf{U}_f\sigma'(\textbf{s}_f)\odot\textbf{c}_t+\textbf{U}_i\sigma'(\textbf{s}_i)\odot\tanh(\textbf{s}_c)
	+\sigma(\textbf{s}_i)\odot(\textbf{U}_g\tanh'(\textbf{s}_c)) \numberthis \label{eq:ctht}
\end{align*}
would typically take values of zero because $\sigma(\textbf{s}_i) $, $ \sigma'(\textbf{s}_f)$ and $\sigma'(\textbf{s}_i)$ are centred near zero, but $ \pdydx{\textbf{h}_{t+1}}{\textbf{h}_{t}} $ depends on gradients w.r.t. the output gate and new-input functions ($ \textbf{o}_{t+1} $ and $ \tilde{\textbf{c}}_{t+1} $), resulting in a summation of non-zero gradients. Initializing the biases of these two gates such that $ \pdydx{J}{\textbf{h}_t}\pdydx{\textbf{h}_t}{\textbf{c}_t}=0 $ could provide a better solution for the LSTM and we leave exploration of this for future work.

In practice the gradients are not as ill-conditioned as we have described here because the gate activations are not homogeneous; some gate-cell combinations track short-term dependencies and others track long-term dependencies. However, with all the initializations kept the same, these derivations could explain why the JANET could be easier to train than the LSTM.
%Moreover, the chrono initializer pertains to the biases of the network and is clearly beneficial to the JANET; comparing eq. \ref{eq:janet_deriv} and \ref{eq:ctht}, the JANET allows a direct dependency of the gradient on the bias of the forget gate (via the $ \sigma(\textbf{s}_f) $ term in eq. \ref{eq:janet_deriv} ), whereas the bias dependencies are not as direct for the LSTM. 

We have shown how the simplification of the LSTM could lead to a better-conditioned training regime, we follow with the theoretical computational savings gleaned by this simplification.

\subsection{Theoretic computational benefits}
Hardware efficient machine learning is a field of study by itself \citep{AdolfFathomreferenceworkloads2016,HintonDistillingKnowledgeNeural2015,SindhwaniStructuredTransformsSmallFootprint2015,HanDeepCompressionCompressing2015,WangAcceleratingRecurrentNeural2017}. The general aim is to maintain the same level of accuracy but require less computational resource in the process. Usually, this applies to only the forward pass efficiency of the network, i.e., being able to run a trained network on a small device. This is the same goal we have for our simplified version of the LSTM. If we assume the accuracies of the JANET and the LSTM to be the same, how much do we save on computation?

Consider an $ n_1 \times n_2 $ LSTM layer that has $ n_1 $ inputs and $ n_2 $ hidden units, then we have $ \textbf{x}_t \in\mathbb{R}^{n_1},\ \{\textbf{c}_t\ ,\ \textbf{h}_t\ ,\ \textbf{b}_j\}\in\mathbb{R}^{n_2},\ \textbf{W}_j\in\mathbb{R}^{n_1\times n_2}, \textbf{U}_j\in\mathbb{R}^{n_1\times n_2}$. For the LSTM we have $ j=\{i,o,f,c\} $, and the total number of parameters is $ 4(n_1n_2 + n_2^2 + n_2) $. For the JANET we have $ j=\{f,c\} $, and the total number of parameters is $ 2(n_1n_2 + n_2^2 + n_2) $. Thus we reduce the number of parameters by half, but what does this mean in terms of memory consumption and computational cost? A proxy for the required memory is the number of values that need to be in memory at each step; e.g., the LSTM requires $ n_1+n_2+n_2+4(n_1n_2 + n_2^2 + n_2) $ values to be stored. Since this value is dominated by the $ 4\times n_2^2 $ term (typically a hidden state size $ n_2 \geq 100$ is used), the JANET would require approximately half of the memory required by an LSTM in a forward pass. \citet{AdolfFathomreferenceworkloads2016} showed that matrix and element-wise multiplication operations each constitute roughly half of the computation required by an LSTM. With the JANET, the processing required for element-wise multiplications is reduced by one third because there are no output gate element-wise multiplications. Thus, the total processing power required by the JANET is roughly $ 0.5+\frac{2}{3}\times 0.5 = \frac{5}{6}^\mathrm{ths} $ of the processing power required by the LSTM.

If we assume that the electrical power consumed by the memory component of our device is 5\% of that consumed by the processor \citep{AcarCPUConsideringmemory2016}, then the JANET will consume approximately $ 0.95\times \frac{5}{6} + 0.05\times0.5 = 0.817$ of the electrical power consumed by the LSTM. However, this ratio is a theoretical estimation and would be different in practice.

Such computational efficiencies are particularly beneficial when applications involve resource-constrained devices. If our simplification of the LSTM is able to provide the same classification accuracy as the standard LSTM, this would be an essential step towards hardware efficient LSTMs. 

\section{Experiments and results}\label{sec:janet_results}
We start by evaluating the performance of the JANET on three publicly available datasets. These comprise the MNIST, permuted MNIST (pMNIST) \citep{arjovsky2016unitary}, and MIT-BIH arrhythmia datasets. The permuted MNIST dataset is the same as the MNIST dataset, except, the pixels in each image have been permuted in the same random order. As
stated by \citet{arjovsky2016unitary}, the MNIST images have regular distinctive patterns much shorter than the 784-long input sequences; permuting the pixels create longer-term dependencies that are harder for LSTMs to learn.

Single heartbeats were extracted from longer filtered signals on channel 1 of the MIT-BIH dataset \citep{MoodyimpactMITBIHArrhythmia2001,goldberger2000physiobank} by means of the \textit{BioSPPy} package \citep{biosppy}. The signals were filtered using a bandpass FIR filter between 3 and 45~Hz, and the Hamilton QRS detector \citep{HamiltonOpensourceECG2002} was used to detect and segment single heartbeats. We chose the four heartbeat classes that are best represented over different patients in the dataset: normal, right bundle branch block, paced, and premature ventricular contraction. The resulting dataset contained 89,670 heartbeats, each of length 216 time steps, from 47 patients. We randomly split the data over patients to have heartbeats from 33 train-, 5 validation-, and 9 test-patients (70:10:20). An acceptable split was considered to have all classes in each set contain at least $ 0.9\gamma\times $\textit{smallest-class-size} data points, where $ \gamma $ is the split fraction (0.7, 0.1, or 0.2). The standard split was used in the case of MNIST.

For the MNIST dataset we used a model with two hidden layers of 128 units, 
whereas a single layer of 128 units was used for the pMNIST and MIT-BIH 
datasets. All the networks were trained using {\em Adam} 
\citep{KingmaAdamMethodStochastic2015} with a learning rate of 0.001 and a 
minibatch size of 200. Dropout of 0.1 was used on the output of the recurrent 
layers, and a weight decay factor of 1e-5 was used. For the LSTM and the JANET, 
chrono initialization was employed. The models were trained for 100 epochs and 
the best validation loss was used to determine the final model. Furthermore, 
the gradient norm was clipped at a value of 5, and the models were implemented 
using Tensorflow \citep{tensorflow2015-whitepaper}.

In table \ref{tab:janet_results} we present the test set accuracies achieved for the three different datasets. In addition to JANET and the standard LSTM, we show the results obtained with the standard recurrent neural network (RNN) and other recent RNN modifications. The means and standard deviations from 10 independent runs are reported. The code to reproduce these experiments is available online: \url{https://github.com/JosvanderWesthuizen/janet}.

\begin{table}[h]
	\centering
	\caption[Recurrent neural network architecture comparisons]{Accuracies [\%] for different recurrent neural network architectures. All networks have a single hidden layer of 128 units unless otherwise stated. The means and standard deviations from 10 independent runs are presented. The best accuracies of our experiments are presented in bold as well as the best cited results.}
	\begin{threeparttable}
	\begin{tabular}{cccc}
		\toprule
		Model & MNIST & pMNIST & MIT-BIH \\
		\midrule
		JANET &
		\textbf{99.0} $ \pm $ 0.120 &
		\textbf{92.5} $ \pm $ 0.767 &
		\textbf{89.4} $ \pm $ 0.193 \\
		LSTM & 98.5 $ \pm $ 0.183 &
		91.0 $ \pm $ 0.518 &
		87.4 $ \pm $ 0.130 \\
		RNN & 10.8 $ \pm $ 0.689  &
		67.8 $ \pm $ 20.18  &
		73.5 $ \pm $ 4.531 \\
		uRNN \citep{arjovsky2016unitary}
		& 95.1  & 91.4  & - \\
		iRNN   \citep{le_simple_2015}
		& 97.0 & 82.0 & - \\
		tLSTM\tnote{a} \hspace{0.05cm}  \citep{HeWiderDeeperCheaper2017} 
		& \textbf{99.2} & \textbf{94.6}  & - \\
		stanh RNN\tnote{b} \citep{ZhangArchitecturalComplexityMeasures2016}		 & 98.1  & 94.0  & - \\
		\bottomrule
	\end{tabular}
	\begin{tablenotes}
		\footnotesize
		\item[a] Effectively has more layers than the other networks.
		\item[b] Single hidden layer of 95 units.
	\end{tablenotes}
	\end{threeparttable}
	\label{tab:janet_results}
\end{table}

Surprisingly, the results indicate that the JANET yields higher accuracies than the standard, LSTM. Moreover, JANET is among the top performing models on all of the analysed datasets. Thus, by simplifying the LSTM, we not only save on computational cost but also gain in test set accuracy.

As in \citet{ZhangArchitecturalComplexityMeasures2016}, due to the 10 to 20 long subsequences of consecutive zeros (see section \ref{sec:janet}), we found training of LSTMs to be harder on MNIST compared to training on pMNIST. By harder, we mean that gradient problems and bad local minima cause the objective function to have a rougher and consequent slower descent than the smooth monotonic descent experienced when training is easy. This does not mean that achieving near-perfect classification is more difficult; near-perfect classification on MNIST is relatively easy, whereas the longer-range dependencies in the pMNIST dataset render near-perfect classification difficult. This pMNIST permutation, in fact, blends the zeros and ones for each data point, giving rise to more uniform sequences, which make training easier. 

\begin{figure}[h]
	\centering
	\includegraphics[width=0.65\textwidth]{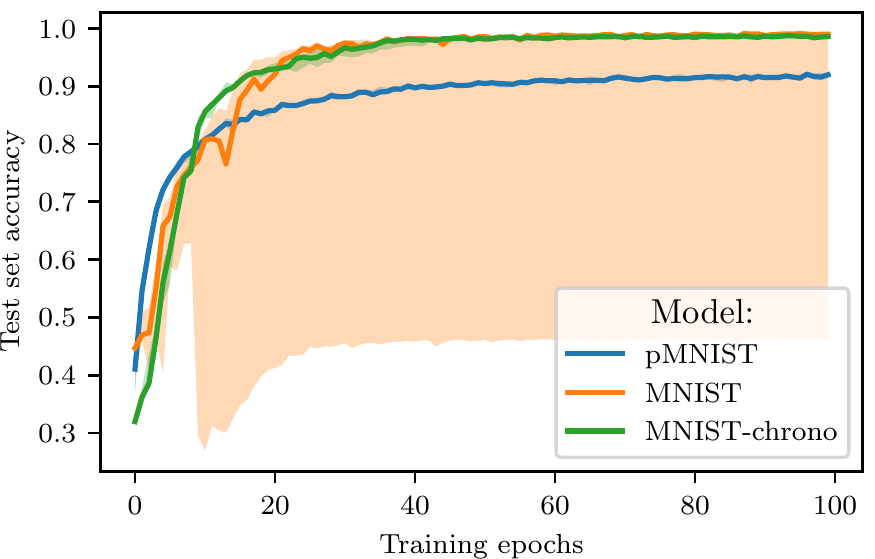}
	\caption[Test accuracies during training for the LSTM on MNIST and pMNIST]{Test accuracies during training for the LSTM on MNIST and pMNIST. The median values are shown with the 10$ ^\mathrm{th} $ and $ 90^\mathrm{th} $ percentiles shaded (for the green and blue curves, the percentiles cover too small an area to see). MNIST-chrono refers to the chrono-initialized LSTM. There was no discernible difference between a chrono-initialized and standard-initialized LSTM on pMNIST. The plots indicate that training is harder on the MNIST dataset, with both LSTM models having a rougher and slower ascent to the optimum accuracy than the model trained on pMNIST. Furthermore, the standard-initialized LSTM catastrophically failed for one of the 10 simulations.}
	\label{fig:mnist_vs_pmnist}
\end{figure}

In figure \ref{fig:mnist_vs_pmnist} we elucidate the difficulty of training on MNIST digits, processed in scanline order. We show the median values with the 10$ ^\mathrm{th} $ and $ 90^\mathrm{th} $ percentiles shaded. From the figure, LSTMs clearly have a rougher ascent in accuracy on MNIST than on pMNIST and can sometimes fail catastrophically on MNIST. The chrono initializer prevents this catastrophic failure during training, but it results in a lower optimum accuracy. On the pMNIST dataset, there were no discernible differences between the chrono and standard-initialized LSTMs -- the benefits of chrono initialization for LSTMs are not obvious on these datasets. 

\begin{figure}[h]
	\centering
	\includegraphics[width=0.7\textwidth]{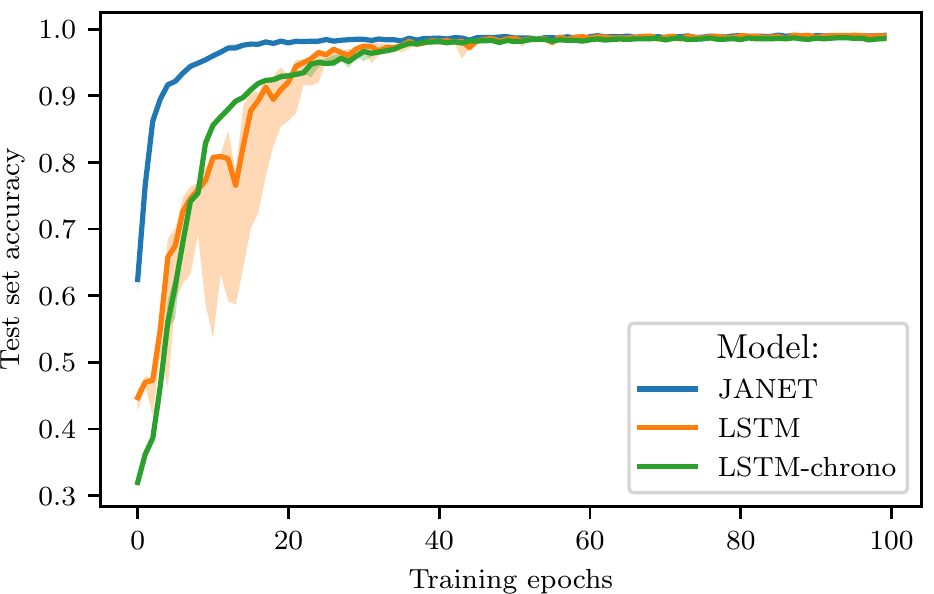}
	\caption[Comparing test set accuracies over training epochs for the JANET and the LSTM]{Comparing test set accuracies over training epochs for the JANET and the LSTM on MNIST. The median values are plotted with the 25$ ^\mathrm{th} $ and $ 75^\mathrm{th} $ percentiles shaded (for the green and blue curves, the percentiles cover too small an area to see). LSTM-chrono refers to the LSTM that is chrono-initialized. Compared with the LSTM, the JANET has a quicker and smoother ascent of test set accuracy during training.}
	\label{fig:janetvlstm}
\end{figure}

As described in section \ref{sec:janet}, the JANET allows skip connections over time steps of the sequence. In figure \ref{fig:janetvlstm} we show how these skip connections result in the JANET being more efficient to train than the LSTM on the MNIST dataset. The median values of the test set accuracies during training are plotted, with the 25$ ^\mathrm{th} $ and $ 75^\mathrm{th} $ percentiles shaded. There is a recent machine learning theme of creating models that are easier to optimize instead of creating better optimizers, which is difficult \citep[\S10.11]{GoodfellowDeepLearning2016}. Being an easier to train version of the LSTM, the JANET continues this theme.

\begin{figure}[h]
	\centering
	\includegraphics[width=0.7\textwidth]{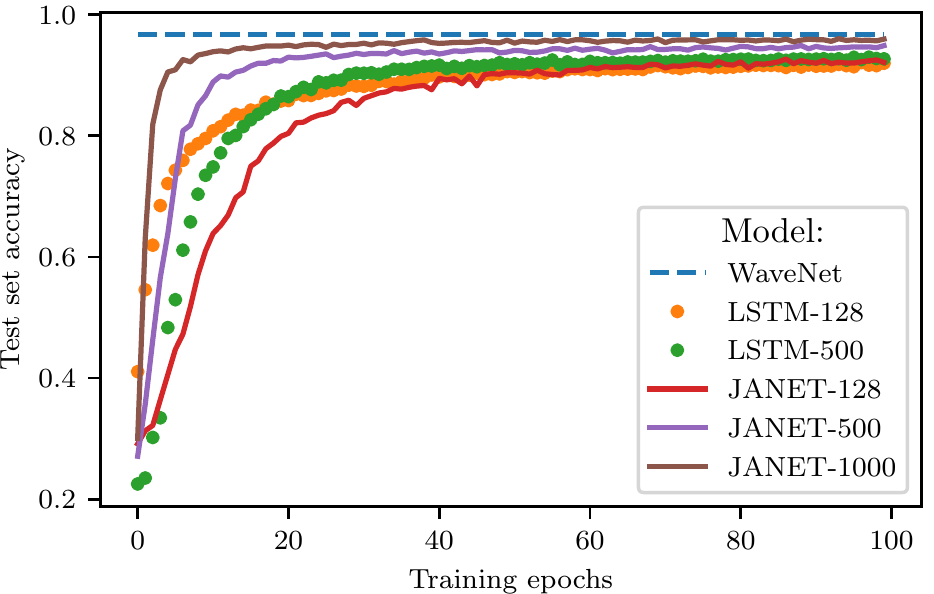}
	\caption[pMNIST accuracy with increased layer sizes]{The accuracy achieved on pMNIST for different layer sizes of the JANET and the LSTM. Median values are shown with the 10$ ^\mathrm{th} $ and $ 90^\mathrm{th} $ percentiles shaded. Both networks were chrono-initialized and the sizes of their single hidden layer are indicated in the legend. The dashed blue depicts the best-reported accuracy on pMNIST \citep{HeWiderDeeperCheaper2017}, which was achieved by a WaveNet \citep{VanDenOordWaveNetGenerativeModel2016}. The JANET clearly improves with a larger layer and performs almost as well as the WaveNet.
		}
	\label{fig:large_janet}
\end{figure}

Given the success of the JANET on the pMNIST dataset (table \ref{tab:janet_results}), we experimented with larger layer sizes. In figure \ref{fig:large_janet} we illustrate the test set accuracies during training for different layer sizes of the LSTM and the JANET. Additionally, we depict the best-reported accuracy on pMNIST \citep{HeWiderDeeperCheaper2017} by the dashed blue line. This best accuracy of 96.7\% was achieved by a WaveNet \citep{VanDenOordWaveNetGenerativeModel2016}, a network with dilated convolutional neural network layers. The dilation increases exponentially across the layers and essentially enables a skip connection mechanism over multiple time steps.

The results show that the JANET not only outperforms the LSTM, but it competes with one of the best performing models on this dataset. With 1000 units in a single hidden layer the JANET achieves a mean classification accuracy of 95.0\% over 10 independent runs with a standard deviation of 0.48\%. The benefit of more units is unclear for the LSTM, which has a similar performance with 500 and 128 units to that of the JANET with 128 units. Furthermore, our models were trained on a Nvidia GeForce GTX 1080 GPU, and the largest LSTM we could train was an LSTM with 500 units. Even with a minibatch size of 1, the LSTM with 1000 units was too large to fit into the 8Gb of GPU memory.

Note that the WaveNet performed worse than the JANET on the standard MNIST dataset, achieving a classification accuracy of 98.3\% compared to the JANET's 99.0\%. The WaveNet results presented here were produced by \citet{ChangDilatedRecurrentNeural2017} using 10 layers of 50 units each. The WaveNet gains additional skip connections with more layers, the JANET gains additional skip connections with more units per layer.

To further ensure that the JANET performs at least as well as the LSTM, we compare the models on two commonly used synthetic tasks for RNN benchmarks. These are known as the copy task and the add task \citep{arjovsky2016unitary,TallecCanrecurrentneural2018,hochreiter_long_1997}.

\paragraph{Copy task} Consider 10 categories $ \{a_i\}_{i=0}^9 $. The input takes the form of a $ T+20 $ length sequence of categories. The first 10 entries, a sequence that needs to be remembered, are sampled uniformly, independently, and with replacement from $ \{a_i\}_{i=0}^7 $. The following $ T-1 $ entries are $ a_8 $, a dummy value. The next single entry is $ a_9 $, representing a delimiter, which should indicate to the model that it is now required to reproduce the initial 10 categories in the output sequence. Thus, the target sequence is $ T+10 $ entries of $ a_8 $, followed by the first 10 elements of the input sequence in the same order. The aim is to minimize the average cross entropy of category predictions at each time step of the sequence. This translates to remembering the categorical sequence of length 10 for $ T $ time steps. The best that a memoryless model can do on the copy task is to predict at random from among possible characters, yielding a loss of $ \frac{10\log8}{T+20} $ \citep{arjovsky2016unitary} \footnote{The first $ T+10 $ entries are assumed to be $ a_8 $, giving a loss of $ -\frac{1}{T+20}(\sum^{T+10}0 + \sum^{10}\sum^{8}\frac{1}{8}\log\frac{1}{8} )$.}.

\paragraph{Add task} Here each input consists of two sequences of length $ T $. 
The first sequence consists of numbers sampled at random from $ 
\mathcal{U}[0,1] $. The second sequence, with exactly two entries of one and 
the remainder zero, is an indicator sequence. The first 1 entry is located 
uniformly at random within the first half of the sequence, and the second is 
located uniformly at random in the second half of the sequence. The scalar 
output corresponds to the sum of the two entries in the first sequence 
corresponding to the non-zero entries of the second sequence. A naive strategy 
would be to predict a sum of 1 regardless of the input sequence, which would 
yield a mean squared error of 0.167, the variance of the sum of two independent 
uniform distributions \citep{arjovsky2016unitary}.

\begin{figure}[h]
	\centering
	\input{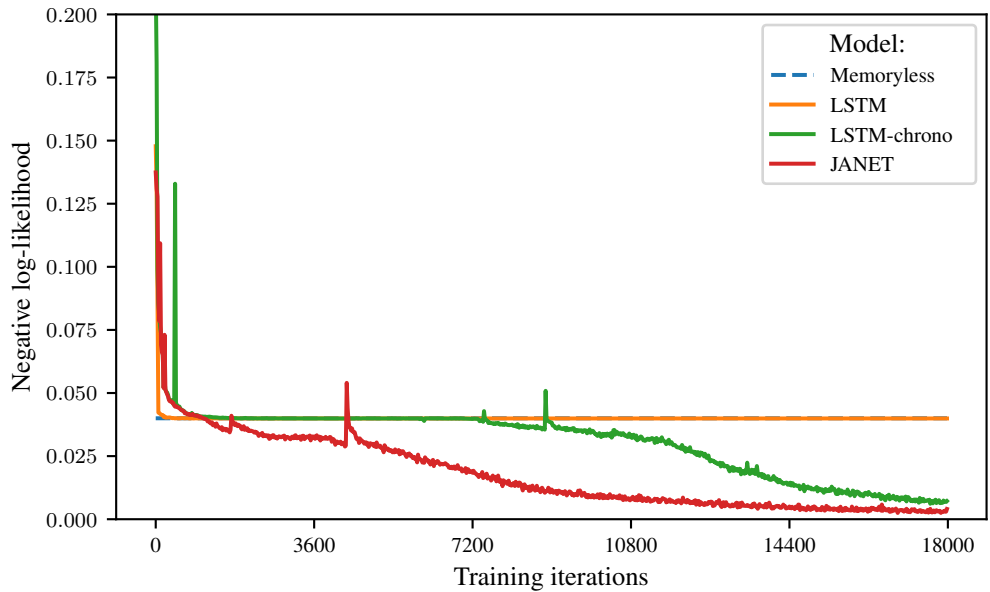}
	\caption[Comparing the JANET and the LSTM on the copy task]{Copy task -- comparing the negative log-likelihood of the JANET and the LSTM on the copy task with $ T=500 $ -- lower is better. The LSTM without chrono initialization performs the same as the memoryless baseline, the same as the results in \citet{arjovsky2016unitary}. Compared to the chrono-initialized LSTM, the JANET converges faster and to a better optimum. }
	\label{fig:copy}
\end{figure}

\begin{figure}[H]
	\centering
	\includegraphics[width=\textwidth]{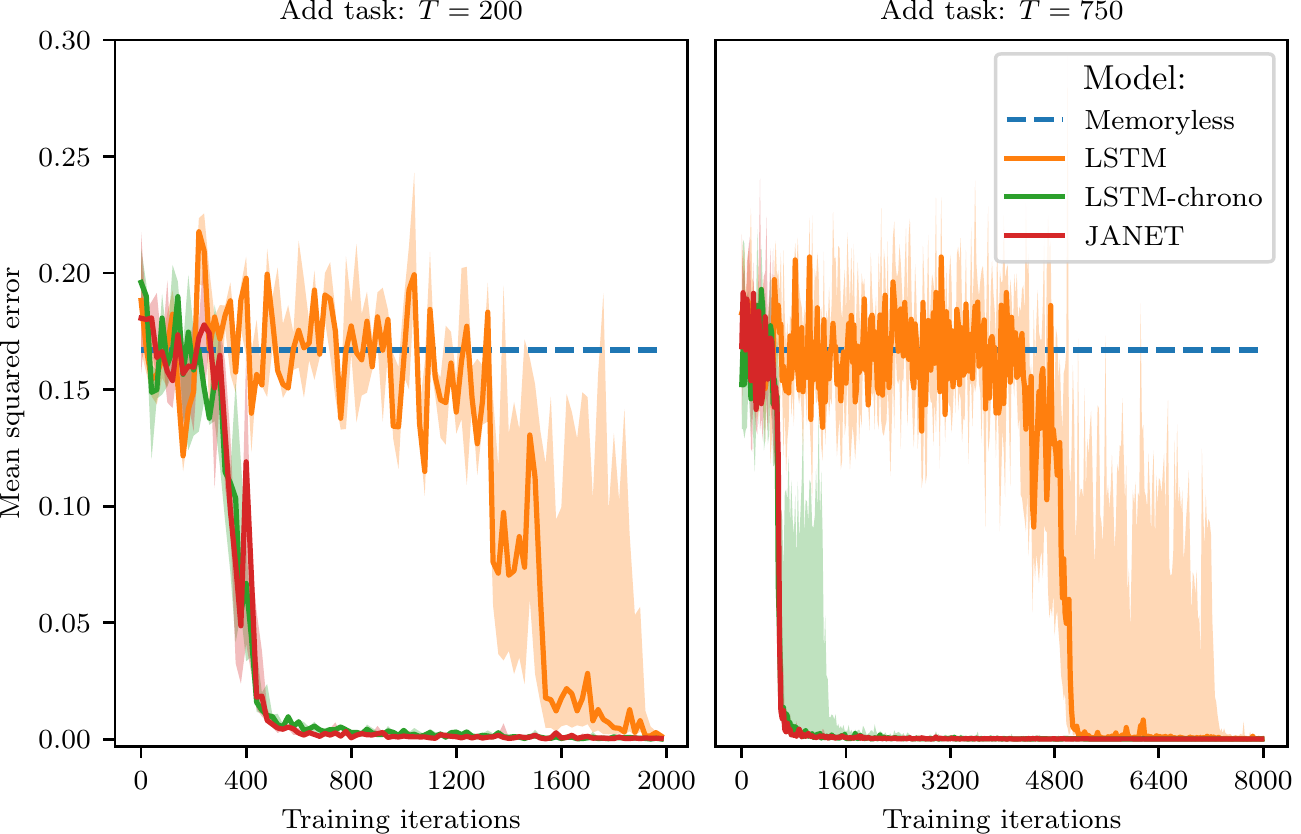}
	\caption[Comparing the JANET and the LSTM on the add task]{Add task-- comparing the mean squared error of the JANET and the LSTM on the add task -- lower is better. The median values of 10 independent runs are shown with the 10$ ^\mathrm{th} $ and $ 90^\mathrm{th} $ percentiles shaded for the add task with $ T=200 $ (\textit{left}) and $ T=500 $ (\textit{right}). Both graphs are displayed with the same y-scale. In both tasks, the standard-initialized LSTM performs the worst. The JANET performs as well as the chrono-initialized LSTM, and slightly better when $ T=750 $.}
	\label{fig:add}
\end{figure}

We follow \citet{TallecCanrecurrentneural2018} and use identical hyperparameters for all our models with a single hidden layer of 128 units. The models were trained using {\em Adam} with a learning rate of 0.001 and a minibatch size of 50. We illustrate the results for the copy task with $ T=500 $, the maximum sequence length used in \citep{arjovsky2016unitary}, in figure \ref{fig:copy}. For the addition task, we explored values of 200 and 750 for $ T $; the results are presented in figure \ref{fig:add}.

In both tasks, we achieve similar results to those reported by \citet{TallecCanrecurrentneural2018} and \citet{arjovsky2016unitary}, and the standard-initialized LSTM performs the worst among the three techniques. Compared to the chrono-initialized LSTM, the JANET converges faster and to a better optimum on the copy task. On the add task, the chrono-initialized LSTM and the JANET have a similar performance, with the latter being slightly better for larger $ T $. The copy task is arguably more memory intensive than the add task. This could explain why the JANET, which has built-in long-term memory capability, would outperform the LSTM on the copy task.

\section{Discussion and conclusion}
In this work, we proposed a simplification of the LSTM that employs only the forget gate and uses chrono-initialized biases. The proposed model was shown to achieve better generalization than the LSTM on synthetic memory tasks and on the MNIST, pMNIST, and MIT-BIH arrhythmia datasets. Additionally, the model requires half of the number of parameters required by an LSTM and two-thirds of the element-wise multiplications, permitting computational savings. The JANET is well-suited for applications to continuous0valued time series with long-term memory requirements. For example, medical time series often have an outcome after several time steps and could have sections of consecutive zero-valued entries. We expect the LSTM to outperform the JANET on next-word prediction tasks where inputs are discrete and non-zero, and predictions are made at each time step.

The unreasonable effectiveness of the proposed model could be attributed to the 
combination of fewer nonlinearities and chrono initialization. This combination 
enables skip connections over entries in the input sequence. As described in 
section \ref{sec:janet}, the skip connections created by the long-range cells 
allow information to flow unimpeded from the elements at the start of the 
sequence to memory cells at the end of the sequence. For the standard LSTM, 
these skip connections are less apparent and an unimpeded propagation of 
information is unlikely due to the multiple possible transformations at each 
time step. 

Modern neural networks move towards the use of more linear transformations 
\citep[\S8.7.5]{GoodfellowDeepLearning2016}. These make optimization easier by 
making the model differentiable almost everywhere, and by making these 
gradients have a significant slope almost everywhere, unlike the sigmoid 
nonlinearity. Effectively, information is able to flow through many more layers 
provided that the Jacobian of the linear transformation has reasonable singular 
values. Linear functions consistently increase in a single direction, so even 
if the model's output is far from correct, it is clear, simply from computing 
the gradient, which direction its output should move towards to reduce the loss 
function. In other words, modern neural networks have been designed so that 
their {\em local} gradient information corresponds reasonably well to moving 
towards a distant solution; a property also induced by skip connections. What 
this means for the LSTM, is that, although the additional gates should provide 
it with more flexibility than our model, the highly nonlinear nature of the 
LSTM makes this flexibility difficult to utilize and so potentially of little use.\todo{add a conclusion tone that says it is interesting to see how effective only the forget gate is. We show some gradients and some initializers, but ultimately, we need better models}

With some success, many studies have proposed models more complex than the LSTM. This has made it easy, however, to overlook a simplification that also improves the LSTM. The JANET provides a network that is easier to optimize and therefore achieves better results. Much of this work showcased how important parameter initialization is for neural networks. In future work, improved initialization schemes could allow the standard LSTM to surpass the models described in this study.

\section*{Acknowledgements}
We thank Jos{\'e} Miguel Hern{\'a}ndez-Lobato for helpful discussions. This 
work is supported by the Skye Cambridge Trust.

\bibliographystyle{myapalike}
\bibliography{references/My_Library,references/references}

\end{document}